 \documentclass[tablecaption=bottom,wcp]{jmlr} 

\usepackage{booktabs}
\usepackage[load-configurations=version-1]{siunitx} 

\theorembodyfont{\upshape}
\theoremheaderfont{\scshape}
\theorempostheader{:}
\theoremsep{\newline}

\usepackage{caption}
\usepackage{subcaption}
\usepackage{listings, chngcntr, float, amsmath, amssymb, cmap, graphicx, xcolor}
\usepackage{amsmath, multirow}

 \jmlrvolume{60}
 \jmlryear{2017}
 \jmlrworkshop{Conformal and Probabilistic Prediction and Applications}
 \jmlrproceedings{PMLR}{Proceedings of Machine Learning Research}

\title[Inductive Conformal Martingales for Change-Point Detection]{Inductive Conformal Martingales for Change-Point Detection}

 \author{\Name{Denis Volkhonskiy} \Email{dvolkhonskiy@gmail.com}\\
 \addr Yandex School of Data Analysis, Moscow, Russia\\
 \addr Skolkovo Institute of Science and Technology,  Skolkovo, Moscow Region, Russia\\
 \addr Institute for Information Transmission Problems, Moscow, Russia
 \AND
 \Name{Evgeny Burnaev}\Email{e.burnaev@skoltech.ru}\\
 \addr Skolkovo Institute of Science and Technology,  Skolkovo, Moscow Region, Russia\\
 \addr Institute for Information Transmission Problems, Moscow, Russia
 \AND
 \Name{Ilia Nouretdinov} \Email{i.r.nouretdinov@rhul.ac.uk}\\
 \addr Information Security Group and Computer Learning Research Center, Department of Computer Science, Royal Holloway, University of London, London, UK
 \AND
\Name{Alexander Gammerman} \Email{a.gammerman@rhul.ac.uk}\\
 \Name{Vladimir Vovk} \Email{v.vovk@rhul.ac.uk}\\
 \addr Computer Learning Research Center, Department of Computer Science, Royal Holloway, University of London, London, UK
 }

\editors{Alex Gammerman, Vladimir Vovk, Zhiyuan Luo, and Harris Papadopoulos}

\newcommand{\E}[2][]{%
    \ifthenelse{\equal{#2}{}}{\mathbb{E}\left[\,#1\,\right]}{\mathbb{E}_{#1}\left[\,#2\,\right]}%
}

\begin{document}
\hspace{\parindent}

\maketitle

\begin{abstract}
We consider the problem of quickest change-point detection in data streams. Classical change-point detection procedures, such as CUSUM, Shiryaev-Roberts and Posterior Probability statistics, are optimal only if the change-point model is known, which is an unrealistic assumption in typical applied problems. Instead we propose a new method for change-point detection based on Inductive Conformal Martingales, which requires only the independence and identical distribution of observations. We compare the proposed approach to standard methods, as well as to change-point detection oracles, which model a typical practical situation when we have only imprecise (albeit parametric) information about pre- and post-change data distributions. Results of comparison provide evidence that change-point detection based on Inductive Conformal Martingales is an efficient tool, capable to work under quite general conditions unlike traditional approaches.
\end{abstract}


\section{Introduction}

Conformal Martingales (martingales based on the conformal prediction  framework, see \citealt[Section 7.1]{alrw}) are known as a valid tool for testing the exchangeability and i.i.d.\ assumptions. They were proposed in \citet{iid1} and later generalized in \citet{iid2}.

One of rather widespread examples of non-i.i.d.\ data is data with Change-Points (CPs) (see \citealt{basseville1993detection, tartakovsky2014sequential}): we assume an on-line scheme of observations, such that before some moment of time (change-point) observations are i.i.d., and after it observations are also i.i.d., but with some other distribution. Thus, overall observations are not i.i.d. This is the reason why application of Conformal Martingales (CMs) to CP detection is possible.

CP detection problems span many applied areas and include automatic video surveillance based on motion features \citep{Pham2014}, intrusion detection in computer networks \citep{Tartakovsky2006}, anomaly detection in data transmission networks \citep{Casas2010}, anomaly detection for malicious activity \citep{resampling, modsel, malware}, change-point detection in software-intensive systems \citep{quasi, ensembles, longrange},
fault detection in vehicle control systems \citep{Malladi1999,alestra},
detection of onset of an epidemic \citep{macneill1995change},
drinking water monitoring \citep{Guepie2012} and many others. 

Standard statistics for change-point detections, such as Cumulative Sum (CUSUM, \citealt{pagecusum}) and Shiryaev-Roberts (S-R, \citealt{shiryaev1963optimum}, \citealt{roberts1966comparison}), have very strong assumptions about data distributions both before and after the change-point. Usually in practice we do not know the change-point model. 

The first attempt to use CMs for change-point detection was made in \citet{ho2005martingale}. 
However, only two different martingale tests were considered for CP detection. 

CM is defined by two main components: a conformity measure (CM) and a betting function \citep{iid1,iid2}. 
Nowadays there exist different approaches to define conformity measures and betting functions. Thus, a whole zoo of CMs for CP detection can be constructed.

Therefore, the goal of our work is to 
\begin{itemize}
\item propose different versions of CMs for CP detection, based on available as well as newly designed conformity measures and betting functions, specially tailored for CP detection;
\item perform extensive comparison of these CMs with classical CP detection procedures.
\end{itemize}
As classical CP detection procedures we consider CUSUM, Shiryaev-Roberts and Posterior Probability statistics. Also we perform comparison with CP detection oracles, which model a typical practical situation when we have only imprecise information about pre- and post-change data distributions. CP detection statistics, considered in the comparison, enjoy different information about statistical characteristics of data and CP models. 

Comparison is performed on simulated data, corresponding to a classical CP model \citep{basseville1993detection}:
\begin{enumerate}
\item[(a)] i.i.d.\ Gaussian white noise signal,
\item[(b)] as a CP we consider change in the mean from zero initial level.
\end{enumerate}

The results of our statistical analyses clearly show that in terms of mean time delay until CP detection for the same level of false alarms CMs are comparable with CP detection oracles and are not significantly worse than optimal CP detection statistics (requiring full information about CP model). At the same time, opposed to classical CP detection statistics, CP detection based on CMs is non-parametric and can be applied in the wild without significant parameter tuning both in case of one-dimensional and multi-dimensional data streams.

The paper is organized as follows. In Section \ref{sec2} we describe CMs. In Section \ref{sec3} we consider quickest CP detection problem statement and describe optimal CP detection statistics, as well as CP detection approaches based on CMs, defined by different conformity measures and betting functions. In Section \ref{sec4} we consider CP detection oracles. In Section \ref{sec5} we describe a protocol of experiments and provide results of simulations. We list conclusions in Section \ref{sec6}.

\section{Conformal Martingales}
\label{sec2}
First we describe Conformal Prediction framework \citep{alrw}, which can be regarded as a tool, satisfying some natural properties of validity, for measuring the strangeness of observations.

\subsection{Non-Conformity measures and p-values}

Let us denote by $z_1, \ldots, z_n, \ldots$ a sequence of observations, where each observation is represented as a vector in some vector space. Our first goal is to test whether the new observation $z_n$ fits the previously observed observations $z_1, \ldots, z_{n-1}$. In other words, we want to measure how strange $z_n$ is compared to other observations. For this purpose, we use the Conformal Prediction framework \citep{alrw}. The first step is the definition of a \textit{non-conformity measure}, which is a function
\[
   (z,S) \mapsto A(z,S),
\]
mapping pairs $(z,S)$ consisting of an observation $z$ and a finite multiset $S$ of observations to a real number $A(z,S)$ with the following meaning: the greater this value is, the stranger $z$ is relative to $S$. As a simple example, one can consider the Nearest Neighbors conformity measure, where $A(z,S)$ is the average distance from $z$ to its nearest neighbors in $S$.

The second step in the Conformal Prediction framework is the definition of the p-value for the observation $z_n$:
\begin{equation}
\label{pvaluess}
  p_n = p(z_n, z_{n-1}, \ldots, z_1)
  =
  \frac{\#\{i=1\ldots n: \alpha_i > \alpha_{n}  \} + U\#\{i=1\ldots n: \alpha_i = \alpha_{n}\}}{n},
\end{equation}
where $U$ is a random number in $[0,1]$ independent of $z_1,z_2,\ldots$, and the \emph{non-conformity scores} $\alpha_i$ (including $i=n$) are defined by
\begin{equation}\label{eq:alpha}
  \alpha_i = A(z_i,\{z_1,\ldots,z_{i-1},z_{i+1},\ldots,z_n\}),
\end{equation}
i.e., the p-value for the observation $z_n$ is defined, roughly, as the fraction of observations that have non-conformity scores greater than or equal to the non-conformity score $\alpha_n$. Intuitively the smaller p-value is, the stranger the observation is.

\begin{theorem}
\label{main_theorem}
If observations $z_1, \ldots, z_n, \ldots$ satisfy the i.i.d.\ assumption, the p-values $p_1, p_2, \ldots$ are independent and uniformly distributed in $[0, 1]$.
\end{theorem}

The statement of Theorem \ref{main_theorem} (proved in \citealt{iid1}) provides grounds for CP detection: 
\begin{itemize}
\item observations $z_1, \ldots, z_{\theta-1}\sim f_0(z)$ are i.i.d.;
\item $z_{\theta}, z_{\theta+1}, \ldots\sim f_1(z)$ are also i.i.d.;
\item in the case $\theta = 1$, all the observations are i.i.d., and therefore CMs couldn't be used for detecting a CP;
\item since at $\theta\geq2$ the distribution changes, the corresponding p-values $p_1, p_2, \ldots, p_n,\ldots$ are not i.i.d.\ uniform in $[0,1]$ for $n\geq\theta$. 
\end{itemize}
We use this fact for constructing CMs for CP detection.

\subsection{Definition of Exchangeability Martingales}
Given a sequence of random vectors $z_1, z_2, \ldots$ taking values in some observation space $\mathbb{R}^d$, the joint probability distribution of $z_1, \ldots, z_N$ for a finite $N$ is \textit{exchangeable} if it is invariant under any permutation of these random vectors. The joint distribution of the infinite sequence of random vectors $z_1, z_2, \ldots$ is \emph{exchangeable} if the marginal distribution of $z_1, \ldots, z_N$ is exchangeable for every $N$.  By de Finetti's theorem, every exchangeable distribution is a mixture of power distributions (i.e., distributions under which the sequence $z_1,z_2,\ldots$ is i.i.d.).
    
A \textit{test exchanegeability martingale} is a sequence of non-negative random variables $S_0=1, S_1, S_2,\ldots$ such that
\[
   \mathbb{E}(S_{n+1} | S_{1}, \ldots, S_{n}) = S_{n},
   \quad
   n=0,1,2,\ldots,
\]
where $\mathbb{E}$ is the expectation w.r.t.\ any exchangeable distribution (equaivalently, any power) on observations. According to Ville's inequality \citep{ville1939etude}, in this case
\[
    \mathbb{P}(\exists n:\, S_n \geq C) \leq \frac{1}{C},~\forall C \geq 1
\]
under any exchangeable distribution.
If the final value of the martingale is large, we can reject the i.i.d.\ (equivalently, exchangeability) assumption with the corresponding probability.  In the next section we define a way to transform p-values \eqref{pvaluess} into test exchangeability martingales.  An \emph{exchangeability martingale} is defined similarly but dropping the requirements that $S_0,S_1,\ldots$ should be non-negative and that $S_0=1$.

\subsection{Constructing Exchangeability Martingales from p-values}

Given a sequence of p-values, we consider a martingale of the form 
\begin{equation}
\label{prodeq}
S_n = \prod_{i=1}^{n}g_i(p_i), \,n = 1, 2, \ldots,
\end{equation}
where each $g_i(p_i) = g_i(p_i\mid p_1, \ldots, p_{i-1})$ is a \textit{betting function} required to satisfy the condition $\int_{0}^{1}g_{i}(p) dp = 1$.  We can easily verify the martingale property under any exchangeable distribution:
\begin{align*}
    \mathbb{E}(S_{n+1} | S_0, \ldots, S_n) &= \int_{0}^{1} \left\{\prod_{i=1}^{n}  g_i(p_i) \right\} g_{n+1}(p) dp =\\
 &= \left\{\prod_{i=1}^{n}  g_i(p_i) \right\} \int_{0}^{1}g_{n+1}(p) dp =  \prod_{i=1}^{n}  g_i(p_i) = S_n.
\end{align*}
Test exchangeability martingale of the form \eqref{prodeq} are \emph{conformal martingales}.  (It is interesting whether there are any other test exchangeability martingales apart from the conformal martingales.)

The intuition behind the betting function is the following: we would like to penalize the fact that p-values are not uniformly distributed (cf.\ Theorem \ref{main_theorem}). In Section \ref{sec:betting} we describe several betting functions along with  their advantages and disadvantages.
    
\section{Quickest Change-Point detection}
\label{sec3}
\subsection{Problem statement}
\label{sec:changepoint_problem}
We observe sequentially a series of independent observations whose distribution changes from $f_0(z)$ to $f_1(z)$ at some unknown point $\theta$ in time. Formally, $z_1, z_2, \ldots, z_n, \ldots$ are independent random variables such that $z_1, z_2, \ldots, z_{\theta-1}$ are each distributed according to a distribution $f_0(z)$ and $z_{\theta}, z_{\theta+1}, \ldots$ are each distributed according to a distribution $f_1(z)$, where $1\leq\theta\leq\infty$ is unknown. The objective is to detect that a change has taken place ``as soon as possible'' after its occurrence, subject to a restriction on the rate of false detections.

Historically, the subject of change-point  detection first began to emerge in the 1920-1930's motivated by considerations of quality control. When a process is ``in control,'' observations are distributed according to $f_0(z)$. At an unknown point $\theta$, the process jumps ``out of control'' and subsequent observations are distributed according to $f_1(z)$. We want to raise an alarm ``as soon as possible'' after the process jumps ``out of control''.

Current approaches to change-point detection were initiated by the pioneering work of Page (\citeyear{pagecusum}). In order to detect a change in a normal mean from $\mu_0$ to $\mu_1>\mu_0$ he proposed the following stopping rule $\tau$: stop and declare the process to be ``out of control'' as soon as $C_n - \min_{1\leq k\leq n} C_k$ gets large, where $C_k = \sum_{i=1}^k (z_i-\mu^*)$ and $\mu_0<\mu^*<\mu_1$ is suitably chosen. This and related procedures are known as CUSUM (cumulative sum) procedures (see \citealt{shiryaev2010fifty} for a survey).

There are different approaches how to formalize a restriction on false detections as well as to formalize the objective of detecting a change ``as soon as possible'' after its occurrence. The restriction on false detections is usually formalized either as a rate restriction on stopping rule $\tau$, according to which we stop our observations and declare the process to be ``out of control'', or a probability restriction. The rate restriction is usually formalized by a requirement that $\mathbb{E}(\tau\mid\theta = \infty)\geq T$, the probability restriction is usually formalized by a requirement that $\mathbb{P}(\tau<\theta)\leq\alpha$ for all $\theta$. The objective of detecting a change ``as soon as possible'' after its occurrence is usually formalized in terms of functionals of $\tau-\theta$  \citep{shiryaev2010fifty}.

\subsection{Optimal approaches to Change-Point detection}
\label{optimcpdec}
Let us describe main optimal statistics for CP detection. The main assumption here is that a known probability density of observations $f_0(z)$ changes to another known probability density of observations $f_1(z)$ at some unknown point $\theta$. We denote by 
\begin{equation}
\label{postdistr}
L^{\theta}_n = \prod_{i=1}^{\theta-1}f_0(z_i)\prod_{i=\theta}^nf_1(z_i)
\end{equation}
the likelihood of observations $z_1,\ldots,z_n$ when $\theta\in[1,n]$, and by
\begin{equation}
\label{predistr}
L_n = \prod_{i=1}^{n}f_0(z_i)
\end{equation}
the likelihood of observations $z_1,\ldots,z_n$ without CP.

\citet{shiryaev1963optimum} solved the CP detection problem in a Bayesian framework. As prior on $\theta$ the distribution $\mathrm{Geometric}(p)$ is used, i.e., $p(n) = \mathbb{P}(\theta = n) = p(1-p)^{n-1}$, $n = 1,2,\ldots$\,. A loss function has the form $\mathbb{P}(\tau\leq\theta) + c\mathbb{E}(\tau-\theta)^+$, where $(x)^+ = \max(x,0)$ and $c>0$. Shiryaev showed that it is optimal to stop observations as soon as the posterior probability of a change exceeds a fixed level $h$, i.e., $\tau_{\text{PP}} = \inf\{n:\,\varphi_n\geq h\}$, where 
\begin{equation}
\label{ppstat}
\varphi_n = \log\left[\frac{\sum_{\theta=1}^nL^{\theta}_np(\theta)}{L_n(1-p)^n}\right].
\end{equation}

In the non-Bayesian (minimax) setting of the problem, the objective is to minimize the expected detection delay for some worst-case change-time distribution, subject to a cost or constraint on false alarms. Here the classical optimality result is due to Lorden, Ritov and Moustakides \citep{lorden, moustakides, ritov}. They evaluate the speed of detection by $\sup_{\theta}\mathrm{ess}\sup_{\omega}\mathbb{E}((\tau-\theta+1)^+\mid z_1,\ldots,z_{\theta-1})(\omega)$ under the restriction that the stopping rules $\tau$ must satisfy $\mathbb{E}(\tau\mid\theta = \infty)\geq T$.  In fact from results of Lorden, Ritov and Moustakides it follows that Page's aforementioned stopping rule, which takes the form $\tau_{\textrm{CUSUM}} = \inf\{n:\,\gamma_n\geq h\}$ with
\begin{equation}
\label{cusumstat}
\gamma_n =\max_{\theta\in[1,n]} \log\left[\frac{L^{\theta}_n}{L_n}\right],
\end{equation}
is optimal.

Pollak (\citeyear{pollak5,pollak7}) considered another non-Bayesian setting: the speed of detection is evaluated by $\sup_{1\leq\theta<\infty}\mathbb{E}(\tau-\theta\mid\tau\geq\theta)$ under the same restriction on the stopping rules, i.e. $\tau$ must satisfy $\mathbb{E}(\tau\mid\theta = \infty)\geq T$. Pollak proved that the so-called Shiryaev-Roberts statistics \citep{shiryaev1963optimum,roberts1966comparison} is asymptotically ($T\to\infty$) minimax. The corresponding stopping rule has the form $\tau_{\textrm{S-R}} = \inf\{n:\,\psi_n\geq h\}$ with
\begin{equation}
\label{srstat}
\psi_n =
\log\left[\frac{\sum_{\theta=1}^nL^{\theta}_n}{L_n}\right].
\end{equation}

As usual we select parameter $h$ for stopping moments $\tau_{\mathrm{PP}}$, $\tau_{\mathrm{CUSUM}}$ and $\tau_{\mathrm{S-R}}$ in such a way that these stopping moments fulfill the corresponding restrictions on false detections.

The main disadvantage of statistics \eqref{ppstat}, \eqref{cusumstat} and \eqref{srstat} is that we should have full information about the CP model, in particular, we should know data distributions before and after the CP. In most of practical situations such assumptions are unrealizable.

 \subsection{Adaptation of Conformal Martingales for Change-Point detection problem}
Let us describe a modification of Conformal Martingales tailored for the CP detection problem:
    \begin{itemize}
    \item Instead of CMs we use their computationally efficient modification that we call \textit{Inductive Conformal Martingales} (ICMs). The main difference of ICMs from CMs is that to compute a non-conformity measure we use some fixed initial training set $\{z_{-(m-1)}^*, \ldots, z_0^*\}$, i.e., each time we receive a new observation $z_n$ we compute the non-conformity score according to the formula
\begin{equation*}
  \alpha_i = A(z_i,\{z_{-(m-1)}^*, \ldots, z_0^*\})
\end{equation*}
(cf.\ original CMs where $\alpha_i$ are defined by \eqref{eq:alpha}). Intuitively, we fix some training set and evaluate to which extent new observations are strange w.r.t.\ this training set. This approach allows us to speed up computations without destroying the validity (see also section \ref{sec:validity}): one should not recompute all non-conformity scores at each iteration. Another advantage is the possibility of parallelization in the batch mode, i.e., when we receive observations in batches.
     \item One drawback of the original CMs, from the point of view of the performance measures adopted in this paper, is that in the case of i.i.d.\ observations CMs decrease to almost zero values with time. As a result, since CMs are represented as a product of betting functions (see \eqref{prodeq}), it takes CMs a lot of time to recover from zero to some significant value when ``strange'' observations appear. In order to deal with this problem we introduce 
\begin{equation}
\label{modcm}
C_n = \max \{0, C_{n-1} + \log(g_n(p_n))\},
\quad
n=1,2,\ldots,
\end{equation}
where $C_0=0$, $p_n$ is a p-value, and $g_n$ is a betting function. On each iteration we cut the logarithm of the martingale. This modification performs better in terms of the mean delay until CP detection.
\end{itemize}
    
\begin{figure}[bt]
    \centering
    \includegraphics[width=15cm]{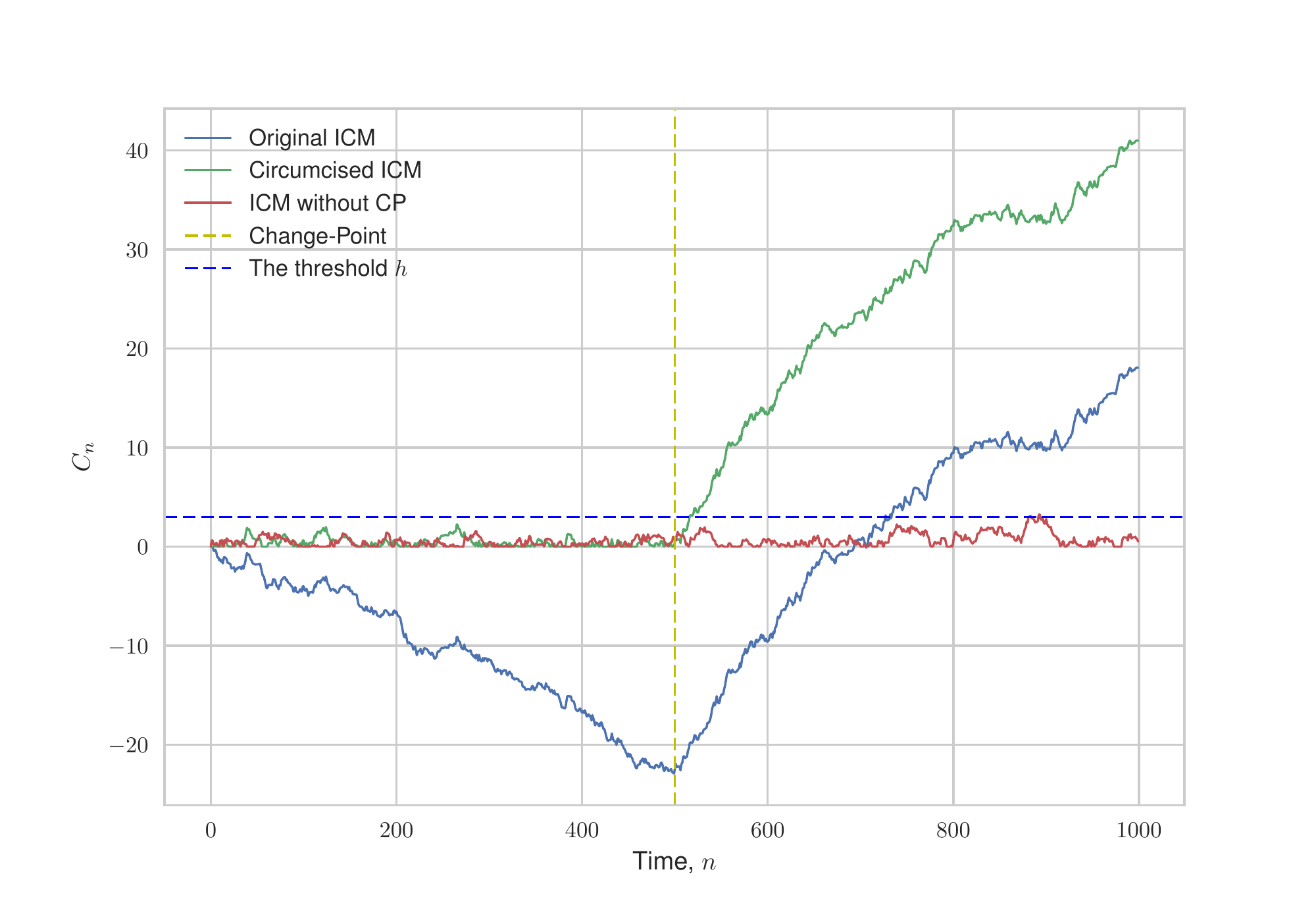}
    \caption{Example of the ICM in case of data with CP (at $\theta=500$) and without CP}
    \label{fig:mart_presentation}
\end{figure}

The complete procedure is summarized in Algorithm \ref{alg:inductive_martingale}. The stopping rule for CP detection has the form $\tau_{\textrm{CM}} = \inf\{n:\,C_n\geq h\}$, where $C_n$ is the modification of the corresponding CM, calculated according to \eqref{modcm}. Notice that
\[
  C_n = \log S_n - \min_{i=1,\ldots,n}\log S_i.
\]
An example of the martingale is given in Fig.\ \ref{fig:mart_presentation}. Here we consider observations from a normal distribution with a unit variance, such that at $\theta=500$ its mean changes from $0$ to $1$. We use two non-conformity measures: $1$ Nearest Neighbor Non-Conformity Measure (1NN NCM) and Likelihood Ratio Non-Conformity Measure (LR NCM), which are described in section \ref{sec:ncm}.
    
        \begin{algorithm}[tb]
    \SetKwInOut{Input}{Input}
    \SetKwInOut{Output}{Output}
    \Input{Training set $\{z_{-(m-1)}^*, \ldots, z_0^*\}$, data $\{z_1,z_2,\ldots\}$, non-conformity measure $A$}
    \Output{Inductive Conformal Martingale $(S_n)_{n\geq1}$ and its modification $(C_n)_{n\geq1}$}
    Randomly shuffle $z_1^*, \ldots, z_m^*$ to induce exchangeability;\\
    Initialize $S_0 = 1$;\\
    \For{$n = 1, 2, \ldots$ } {
      observe new observation $z_n$\\
      calculate non-conformity score $\alpha_n = A(z_n, \{z_{-(m-1)}^*, \ldots, z_0^*\})$\\
      calculate p-value $p_n = \frac{\#\{i=1\ldots n: \alpha_i > \alpha_{n}  \} + U\#\{i=1\ldots n: \alpha_i = \alpha_{n}\}}{n}$, where $U \sim \text{Uniform}[0,1]$\\
      calculate new ICM value $S_n$ according to \eqref{prodeq} and calculate its modification $C_n$ according to \eqref{modcm}, where $g_n(p)$ is a betting function
    }
    \caption{Change-point detection with Inductive Conformal Martingale}
    \label{alg:inductive_martingale}
\end{algorithm}
    
\subsection{Validity}
\label{sec:validity}
\begin{figure}[bt]
    \centering
    \includegraphics[width=15cm]{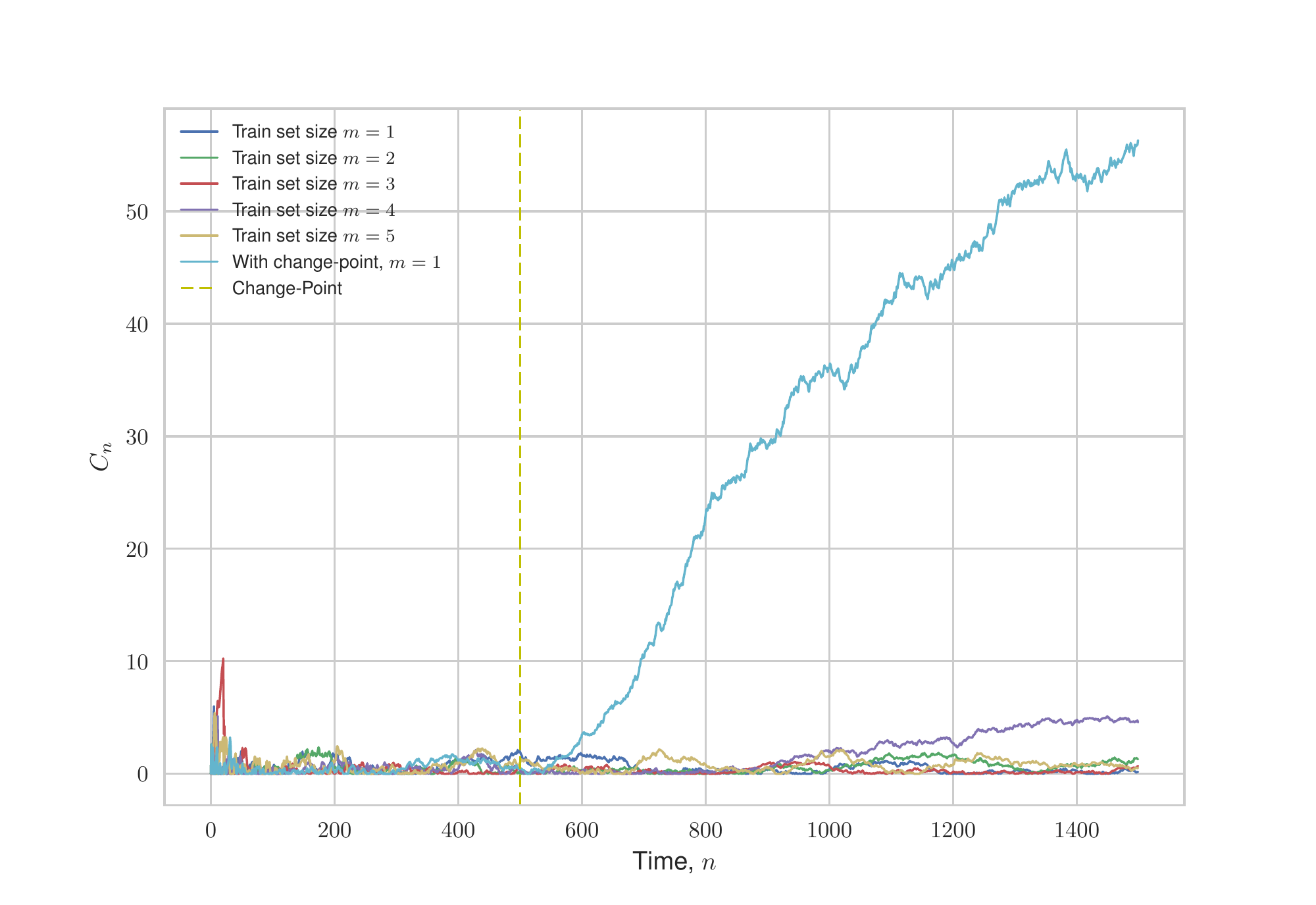}
    \caption{Validity Test of ICM: case of small train sets}
    \label{fig:validity}
\end{figure}

Let us check empirically that our method is valid for small values of train set size $m$ (the theoretical validity is lost because of the transition from $S_n$ to $C_n$). For this purpose we generate observations from $\mathcal{N}(\cdot\mid 0, 1)$ without CP and with CP (mean changes from $0$ to $1$ at $\theta = 500$). Here $\mathcal{N}(z\mid\mu,\sigma^2)$ is a value at point $z$ of a normal density with mean $\mu$ and variance $\sigma^2$. We use $k$ Nearest Neighbor non-conformity measure (see section \ref{sec:ncm} below). We plot ICM for train set sizes $m \in \{1, 2, 3, 4, 5\}$ in Fig.\ \ref{fig:validity}. Results of simulations, provided in Fig.\ \ref{fig:validity}, confirm the validity of our approach.

\subsection{Non-Conformity Measures}
\label{sec:ncm}
Let us describe non-conformity measures that we use:
\begin{itemize}
\item $k$ Nearest Neighbors Non-Conformity Measure (kNN NCM). kNN NCM is computed as the average distance to $k$ nearest neighbors. The advantage of this NCM is that it doesn't depend on any assumptions and can be used in a multi-dimensional case;
\item Likelihood Ratio Non-Conformity Measure (LR NCM). One way or another the classical CP detection algorithms (see section \ref{optimcpdec}) are based on a likelihood ratio. Thus it is worth to consider LR NCM. If we denote by $f_0(z)$ and $f_1(z)$ probability density functions before and after the CP, then a reasonable LR NCM would be
\[
\alpha_n=\frac{f_1(z_n)}{f_0(z_n)}.
\]
However, we rarely know $f_i(z)$, $i = 0,1$, exactly. Thus, we should somehow model this lack of information. We assume that $f_i(z)$, $i = 0,1$, belong to some parametric class of densities, i.e., $f_i(z) = f(z\mid\mathbf{c}_i)$, where $\mathbf{c}_i\in\mathbf{C}$, $i = 0,1$ are vectors of parameters. We estimate the value of $\mathbf{c}_0$ by some $\hat{\mathbf{c}}_0$ using the training set $\{z_{-(m-1)}^*,\ldots,z_0^*\}$. We also impose some prior $r(\mathbf{c}_1)$ on the parameter $\mathbf{c}_1$, i.e., we model the data distribution after the CP by $\overline{f}_1(z)=\int f(z\mid\mathbf{c}_1)r(\mathbf{c}_1)d\mathbf{c}_1$. As a result LR NCM has the form
\begin{equation}
\label{LRNCM}
\alpha_n=\frac{\int f(z_n\mid\mathbf{c}_1)r(\mathbf{c}_1)d\mathbf{c}_1}{f(z_n\mid\hat{\mathbf{c}}_0)}.
\end{equation}
E.g., in the one-dimensional case for  $f(z\mid\mu_1) = \mathcal{N}(z\mid\mu_1,\sigma^2)$ and $r(\mu_1) = \mathcal{N}(z\mid\mu_r,\sigma^2_r)$ we get that
\begin{equation}
\label{LRNCM2}
\alpha_n=\frac{\mathcal{N}(z_n\mid\mu_r,\sigma^2+\sigma^2_r)}{\mathcal{N}(z_n\mid\hat{\mu}_0,\sigma^2)},
\end{equation}
where $\hat{\mu}_0 = \frac{1}{m}\sum_{i=1}^mz_{-m+i}$.
\end{itemize}

\subsection{Betting Functions}
\label{sec:betting}
Let us describe Betting Functions that we use:
\begin{itemize}
\item \textit{Mixture Betting Function} was proposed in the very first work on testing exchangeability \citep{iid1}. It doesn't depend on the  previous p-values and has the form
\[
g(p) = \int_0^1 \varepsilon p^{\varepsilon - 1} d\varepsilon.
\]
\item \textit{Constant Betting Function}. We split the interval $[0, 1]$ into two parts at the point $0.5$. We expect p-values to be small if observations are strange:
\[
g(p) = \begin{cases}
1.5, \text{ if $p \in [0, 0.5)$},\\
0.5, \text{ if $p \in [0.5, 1]$}.
\end{cases}
\]

\item \textit{Kernel Density Betting Function} has the form 
\label{sec:kernel_bet_fun}
\[
g_n(p_n) = K_{p_{n-L}, \ldots, p_{n-1}}(p_n).
\]
Here $K_{p_{n-L}, \ldots, p_{n-1}}(p)$ is a Parzen-Rosenblatt kernel density estimate \citep{kernel_estim} based on the previous p-values $\{p_{n-L}, \ldots, p_{n-1}\}$, $L$ being a window size. We use a Gaussian kernel. Since p-values are in $[0,1]$, then to reduce boundary effects we reflect the p-values to the left of zero and to the right of one, construct the density estimate, crop its support back to $[0,1]$ and normalize. \citet{iid2} prove that such an approach provides an asymptotically better growth rate of the exchangeability martingale than any martingale with a fixed betting function. The corresponding martingale is also called the \textit{plug-in martingale}. Let us note that for quicker CP detection we use not all p-values, but only last $L$ of them: $\{p_{n-L}, \ldots, p_{n-1}\}$. Increasing $L$ usually results in an increase of the mean delay, because after the CP we need to collect more observations to estimate the new distribution of p-values correctly.

\item \textit{Precomputed Kernel Density Betting Function}.
To deal with the problem of high mean delay until CP detection, we propose to estimate the kernel density of p-values before constructing any martingale. For this purpose, we have to learn the betting function using some finite length realization of $z_1,z_2,\ldots,z_n,\ldots$, containing an example of a typical CP, and some training set $\{z^*_{-(m-1)}, \ldots, z_0^*\}$. In other words, the realization should contain some CP with position and intensity resembling those of real CPs (say within the accuracy of order of magnitude) we are going to detect while applying the corresponding CM. Particular values of these parameters are specified in experimental Section \ref{sec5}. We compute p-values using \eqref{pvaluess} as in Algorithm \ref{alg:inductive_martingale}. Using them we construct a kernel estimate of p-values density. Further we assume that for data of the same nature p-values will be distributed in a similar way, so we can use this precomputed kernel density betting function for new data realizations. Thus, thanks to the precomputed estimate we can
\begin{itemize}
\item[---] Detect CP faster;
\item[---] Speed-up computations (we don't need to reconstruct  density of p-values for each position of the sliding window).
\end{itemize}
\end{itemize}

\section{Oracles for Change-Point detection}
\label{sec4}
In the current section we describe Oracles for CP detection that we compare with CP detection based on Conformal Martingales. 
\subsection{Motivation to use Oracles}
First we explain why we need to compare CP detection based on CMs with CP detection Oracles:
\begin{itemize}
\item Classical CP detection statistics are optimal in terms of the mean delay (subject to a restriction on the rate of false detections) if data distributions before and after the CP are known. There is no need for them to learn the distributions $f_0$ and $f_1$ before and after the CP.
\item CMs are designed to solve another problem. As far as their validity is concerned, they assume nothing about the distributions $f_i$, $i = 0,1$. They have to learn the distribution $f_0$ before the CP in order to detect a change.
\item The profound difference between the classical setting and the adaptive setting dealt with in conformal prediction can be seen clearly if instead of the problem of quickest CP detection we consider the related problem of gambling (formalized by constructing a test martingale) against the null hypothesis ($f_0$ in the case of classical statistics and i.i.d.\ in the case of conformal prediction) in the presence of a CP.  In the classical case, the growth rate of the optimal test martingale (likelihood ratio) will be exponential since the null hypothesis is simple, whereas in the i.i.d.\ case after an initial period of nearly exponential growth the growth rate will slow down as we start learning that $f_1$ is much closer to being the data-generating distribution than $f_0$ is.
\item Thus for a fair comparison we should compare CP detection based on CMs not with CP detectors from Section \ref{optimcpdec}, which are optimal under known $f_0$ and $f_1$, but with their modifications (oracles, defined in Section \ref{descriptoracle} below) that have plenty of information about pre- and post-change data distributions, but there is still some uncertainty; the oracles only know the parametric models that $f_0$ and $f_1$ are coming from, and the task of competing with them making only a nonparametric assumption (i.i.d.)\ is challenging but not hopeless.
\end{itemize}

\subsection{Description of Oracles}
\label{descriptoracle}
We assume that $f_i(z)$, $i = 0,1$, belong to some parametric class of densities, i.e., $f_i(z) = f(z\mid\mathbf{c}_i)$, where $\mathbf{c}_i\in\mathbf{C}$, $i = 0,1$ are vectors of parameters. 
We impose the same prior $q(\mathbf{c})$ on the parameters $\mathbf{c}_i$, $i = 0,1$. Thus, instead of likelihood \eqref{predistr} of observations without CP we use 
\begin{equation}
\label{predistr2}
\overline{L}_n = \int\prod_{i=1}^{n}f(z_i\mid\mathbf{c}_0) q(\mathbf{c}_0)d\mathbf{c}_0,
\end{equation}
and instead of likelihood \eqref{postdistr} of observations $z_1,\ldots,z_n$  with CP at $\theta\in[1,n]$ we use
\begin{equation}
\label{postdistr2}
\overline{L}^{\theta}_n = \int\prod_{i=1}^{\theta-1}f_0(z_i\mid\mathbf{c}_0)q(\mathbf{c}_0)d\mathbf{c}_0\cdot\int\prod_{i=\theta}^nf_1(z_i\mid\mathbf{c}_1)q(\mathbf{c}_1)d\mathbf{c}_1.
\end{equation}

Oracles are obtained from optimal statistics \eqref{ppstat}, \eqref{cusumstat} and \eqref{srstat} by using $\overline{L}_n$ from \eqref{predistr2} instead of $L_n$ from \eqref{predistr}, and by using $\overline{L}_{n}^{\theta}$ from \eqref{postdistr2} instead of $L_{n}^{\theta}$ from \eqref{postdistr} (cf. with section 2.4.2.1 and example 2.4.2 in \cite{basseville1993detection}).

Let us consider a one-dimensional example. We set $f(z\mid\mu_i) = \mathcal{N}(z\mid\mu_i,1)$, $i = 1,2$ and $q(\mu) = \mathcal{N}(\mu\mid 0,1)$, and we get that
\begin{align}
\overline{L}_n &= \overline{L}_n(z_1,\ldots,z_n)  = \int_{\mathbb{R}}\prod_{i=1}^n\mathcal{N}(z_i\mid\mu_0,1)\mathcal{N}(\mu_0\mid 0,1)d\mu_0\notag \\
& = \left(\frac{1}{\sqrt{2\pi}}\right)^n\sqrt{\frac{2\pi}{n+1}}\exp\left\{-\frac{n\left[\overline{z^2_{1,n}} - \frac{n}{n+1}(\overline{z_{1,n}})^2\right]}{2}\right\},\label{lnonedim}\\
\overline{L}_n^{\theta} &= \overline{L}_n^{\theta}(z_1,\ldots,z_n)  = \int_{\mathbb{R}}\prod_{i=1}^{\theta-1}\mathcal{N}(z_i\mid\mu_0,1)\mathcal{N}(\mu_0\mid 0,1)d\mu_0\int_{\mathbb{R}}\prod_{i=\theta}^{n}\mathcal{N}(z_i\mid\mu_1,1)\mathcal{N}(\mu_1\mid 0,1)d\mu_1\notag \\
& = \frac{1}{\sqrt{(2\pi)^{n-2}\theta(n-\theta+2)}}\exp\left\{-\frac{n\left[\overline{z^2_{1,n}} - \left\{\frac{(\theta-1)^2}{n\theta}\left(\overline{z_{1,\theta-1}}\right)^2 + \frac{(n-\theta+1)^2}{n(n-\theta+2)}\left(\overline{z_{\theta,n}}\right)^2\right\}\right]}{2}\right\},\label{lnonedim2}
\end{align}
where $\overline{z_{m,n}} = \frac{1}{n-m+1}\sum_{i=m}^nz_i$, $\overline{z^2_{m,n}} = \frac{1}{n-m+1}\sum_{i=m}^nz_i^2$. In such a way we model a situation when the Oracle does not know exact values of $\mu_i$, $i=1,2$.

\section{Experiments}
\label{sec5}
In the current section we describe our experimental setup and provide results of experiments. 

\subsection{Experimental setup}

We consider the following experimental setup:
    \begin{itemize}
    \item We use observations $\{z_{-(m-1)}^*, \ldots, z_{0}^*\}$ as a training set for computation of non-conformity scores. We set $m = 200$ in all experiments.
    \item Observations $z_{-(m-1)}^*, \ldots, z_{0}^*,z_1, \ldots, z_{\theta-1}$ are generated from $f_0(z) \sim \mathcal{N}(z\mid 0, 1)$.
    \item Observations $z_{\theta },z_{\theta+1}, \ldots$ are generated from $f_1(z) \sim \mathcal{N}(z\mid \mu_1, 1)$. We consider $\mu_1 \in \{1, 1.5, 2\}$.
    \end{itemize}

As performance characteristics we use:
\begin{itemize}
    \item Mean delay until CP detection $\mathbb{E}_1(\tau - \theta\mid \tau > \theta)$,
    \item Probability of False Alarm (FA) $ \mathbb{P}_0(\tau \leq \theta)$.
\end{itemize}
In all experiments using Monte-Carlo simulations we estimate dependency of the mean delay $\mathbb{E}_1(\tau - \theta\mid \tau > \theta)$ on the probability of the false alarm $ \mathbb{P}_0(\tau \leq \theta)$.

For the LR NCM in \eqref{LRNCM2} we set $\mu_r = 1$, $\sigma^2 = 1$ and $\sigma^2_r=1$, i.e.,
\[
\alpha_n=\frac{\mathcal{N}(z_n\mid 1,2)}{\mathcal{N}(z_n\mid\hat{\mu}_0,1)},
\]
where $\hat{\mu}_0 = \frac{1}{m}\sum_{i=1}^mz_{-m+i}^*$.

In the case of oracle detectors (see section \ref{descriptoracle}) we use likelihoods from \eqref{lnonedim} and from \eqref{lnonedim2} to obtain Posterior Oracle from the optimal statistics \eqref{ppstat}, CUSUM Oracle from the optimal statistics \eqref{cusumstat} and S-R Oracle from the optimal statistics \eqref{srstat}. When calculating Posterior Probability statistics \eqref{ppstat} and Posterior Oracle we set parameter $p$ of the geometric distribution to $\frac{1}{100}$.

In experiments we consider all possible combinations of different types of Oracles, betting functions from section \ref{sec:betting}, non-conformity measures from section \ref{sec:ncm}, as well as different values of $\mu_1\in\{1,1.5,2\}$ and $\theta\in\{100,200\}$. In the case of kNN NCM we set $k$ to $7$.

\subsection{Refinement of the experimental setup}
\begin{figure}[bt]
    \centering
    \includegraphics[width=16cm]{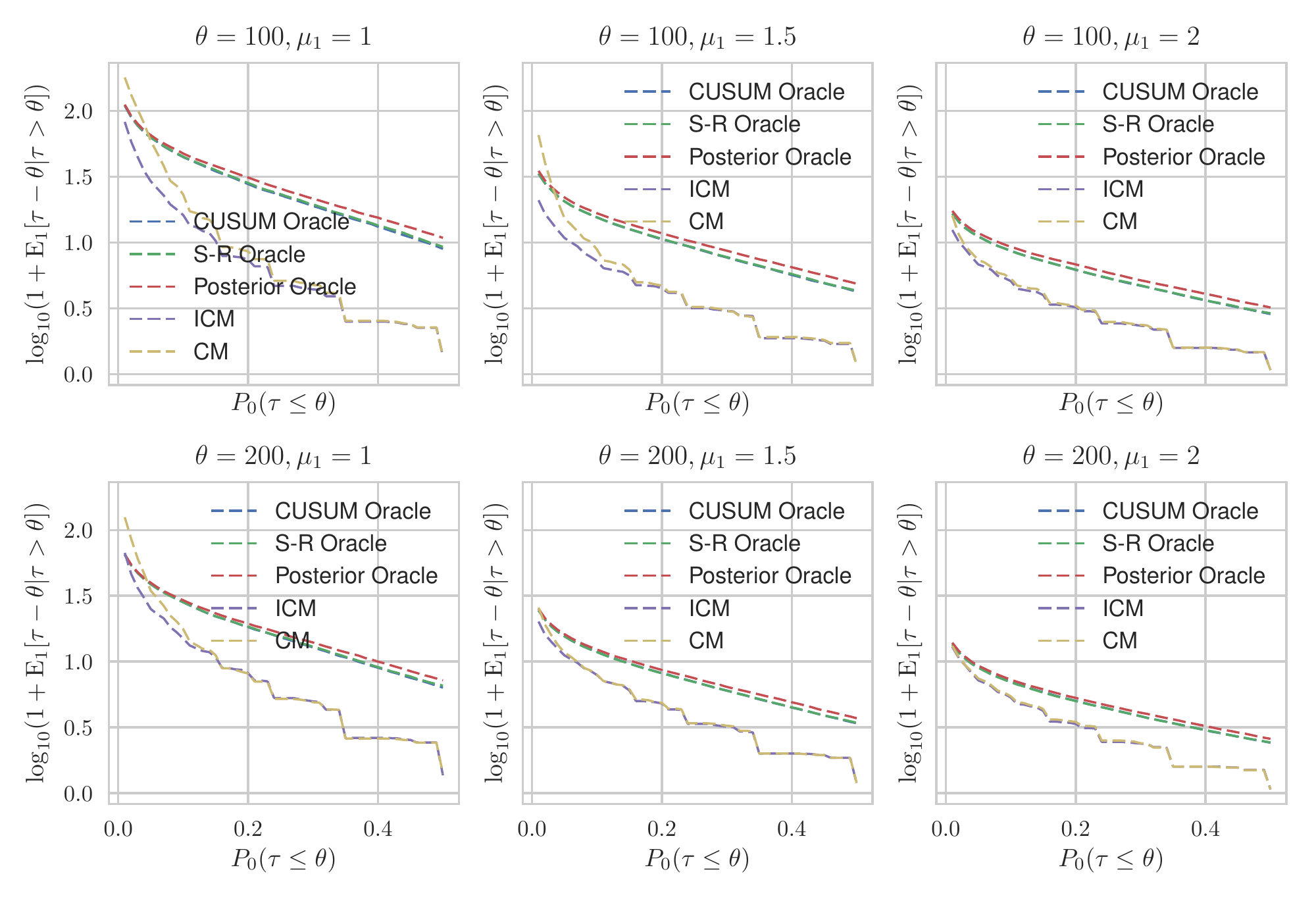}
    \caption{Comparison of ICM and CM for the constant betting function}
    \label{fig:icm_and_cm_const}
\end{figure}

When applying Conformal Martingales both original and inductive versions can be used. First let us check that the inductive version is not worse than the original one. In our comparison we use a simple NCM: $\alpha_i = A(z_i,\{z_1,\ldots,z_{i-1},z_{i+1},\ldots,z_n\} = \left|z_i - \frac{1}{n-1}\sum_{j\neq i}z_j\right|$. In Fig. \ref{fig:icm_and_cm_const} we plot 
estimated dependency of the mean delay on the probability of the false alarm for both ICM and CM with the constant betting function and different oracles. 
As we can see, there is almost no difference in the original and inductive versions. Later we consider only Inductive Conformal Martingales.

\begin{figure}[bt]
    \centering
    \includegraphics[width=16cm]{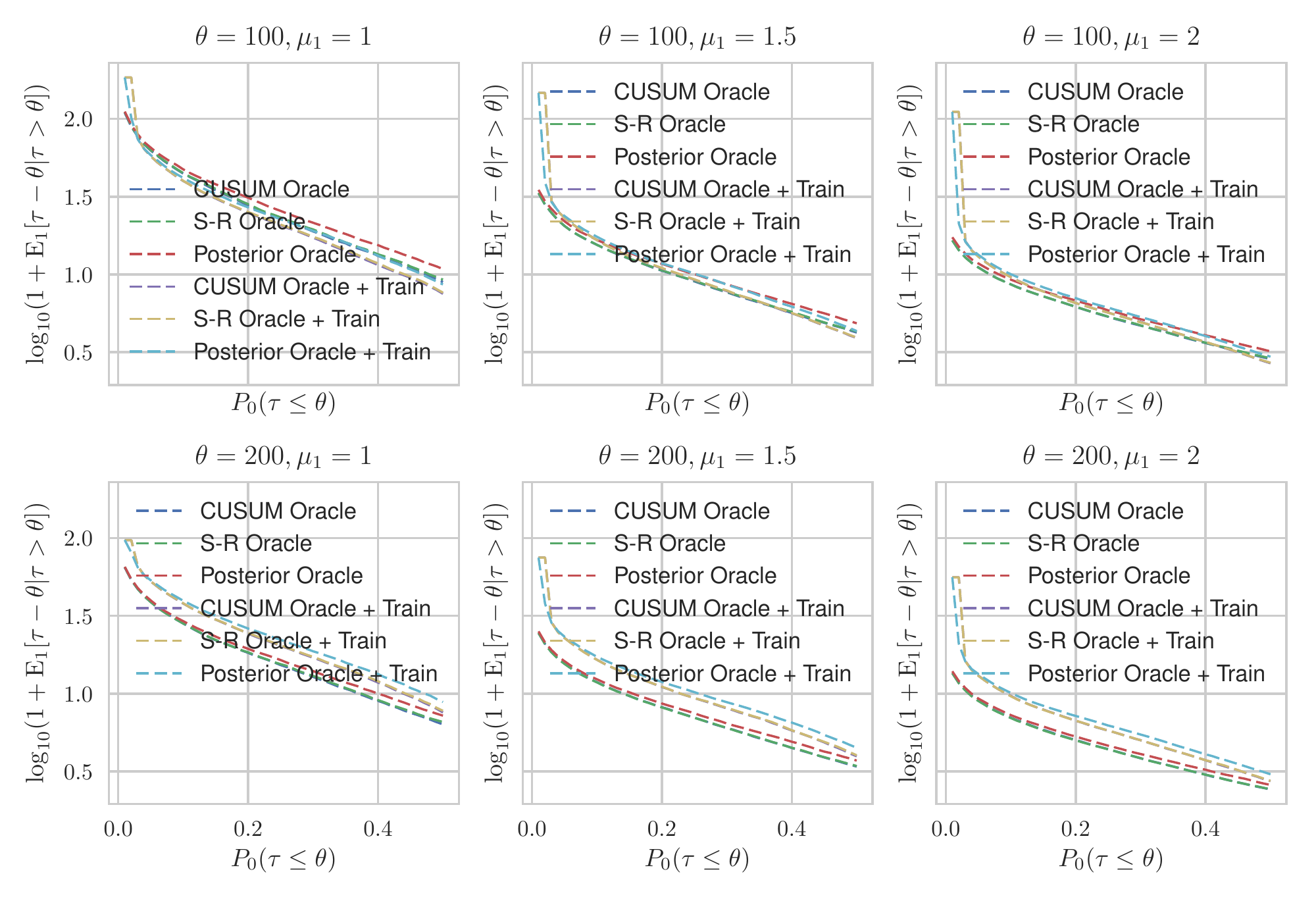}
    \caption{Comparison of Oracles with and without train set}
    \label{fig:long_oracles_20000}
\end{figure}

When calculating the Oracles we can either additionally use the train set $\{z_{-(m-1)}^*, \ldots, z_0^*\}$ or not. Let us check how the addition of the train set influence results. The comparison is presented in Fig. \ref{fig:long_oracles_20000}. We can see that the results are practically the same. Later in the paper when calculating the Oracles we do not use the train set.

\subsection{Constant Betting Function}
\begin{figure}[bt]
    \centering
    \includegraphics[width=16cm]{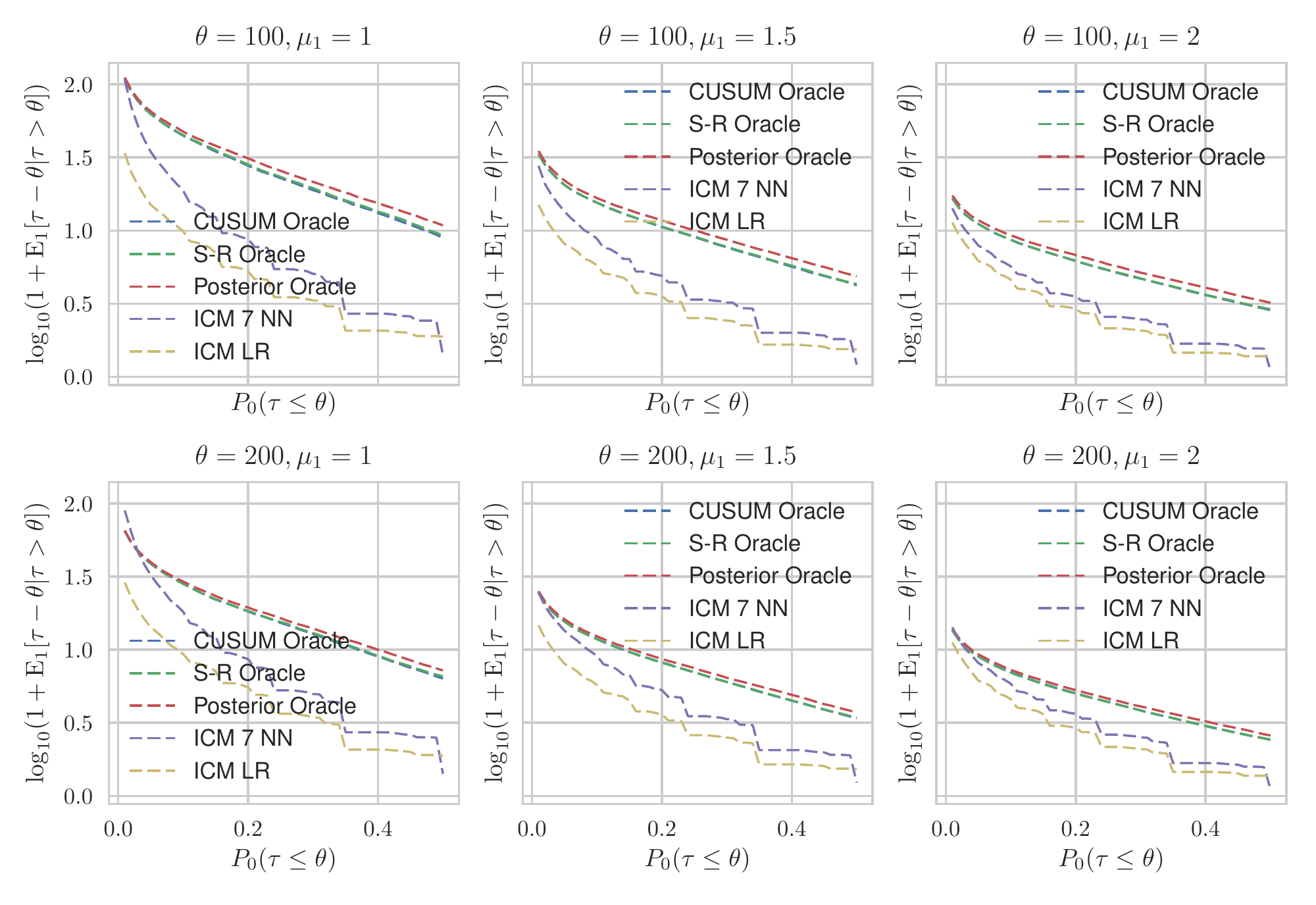}
    \caption{Constant Betting Function}
    \label{fig:const_betting_function}
\end{figure}

Results for Constant Betting Function are in Fig.\ \ref{fig:const_betting_function}. Here \textit{SR} stands for S-R Oracle, \textit{PP} --- for Posterior Oracle, \textit{CUSUM} --- for CUSUM Oracle, \textit{ICM 7 NN} --- for ICM CP detector with $k=7$ nearest neighbor NCM, \textit{ICM LR} --- for ICM CP detector with LR NCM. Mean delays for some values of false alarm probability are in Tab. \ref{tab:const_betting_function}.

\begin{table}[bt]
\centering
\caption{Comparison of ICM (Constant Betting Function) with Oracle by Mean Delay for different False Alarm probabilities}
\label{tab:const_betting_function}
\resizebox{\columnwidth}{!}{%
\begin{tabular}{|c|c|c|c|c|c|c|c|c|c|c|}
\hline
\multicolumn{1}{|l|}{\multirow{2}{*}{\textbf{Param.\textbackslash Probab. of FA}}} & \multicolumn{5}{c|}{\textbf{5\%}}                                                                                                                                                                                                                                                     & \multicolumn{5}{c|}{\textbf{10\%}}                                                                                                                                                                                                                                                    \\ \cline{2-11} 
\multicolumn{1}{|l|}{}                                                             & \textbf{ICM LR}  & \textbf{ICM kNN} & \textbf{\begin{tabular}[c]{@{}c@{}}CUSUM\\ Oracle\end{tabular}} & \multicolumn{1}{c|}{\textbf{\begin{tabular}[c]{@{}c@{}}S-R\\ Oracle\end{tabular}}} & \multicolumn{1}{c|}{\textbf{\begin{tabular}[c]{@{}c@{}}Posterior\\ Oracle\end{tabular}}} & \textbf{ICM LR}  & \textbf{ICM kNN} & \textbf{\begin{tabular}[c]{@{}c@{}}CUSUM\\ Oracle\end{tabular}} & \multicolumn{1}{c|}{\textbf{\begin{tabular}[c]{@{}c@{}}S-R\\ Oracle\end{tabular}}} & \multicolumn{1}{c|}{\textbf{\begin{tabular}[c]{@{}c@{}}Posterior\\ Oracle\end{tabular}}} \\ \hline
$\theta=100, \mu_1=1$ & 14.02 &  33.52 &  61.59 & 62.01 &  64.37 &  8.90 &  17.71 &  43.53 & 43.89 &  46.40 \\ \hline
$\theta=100, \mu_1=1.5$ & 7.08 &  12.51 &  19.51 & 19.51 &  20.98 &  4.79 &  7.79 &  14.50 & 14.51 &  15.67 \\ \hline
$\theta=100, \mu_1=2$ & 5.19 &  6.90 &  10.11 & 10.09 &  10.78 &  3.62 &  4.70 &  7.64 & 7.64 &  8.27 \\ \hline
$\theta=200, \mu_1=1$ & 13.22 &  31.33 &  37.78 & 37.80 &  38.73 &  8.33 &  17.17 &  27.24 & 27.24 &  28.25 \\ \hline
$\theta=200, \mu_1=1.5$ & 7.00 &  12.50 &  14.62 & 14.52 &  15.16 &  4.74 &  8.08 &  10.85 & 10.81 &  11.36 \\ \hline
$\theta=200, \mu_1=2$ & 5.13 &  7.12 &  8.02 & 7.98 &  8.30 &  3.59 &  4.85 &  6.00 & 5.97 &  6.28 \\ \hline
\end{tabular}
}
\end{table}

\subsection{Mixture Betting Function}
\begin{figure}[bt]
    \centering
    \includegraphics[width=16cm]{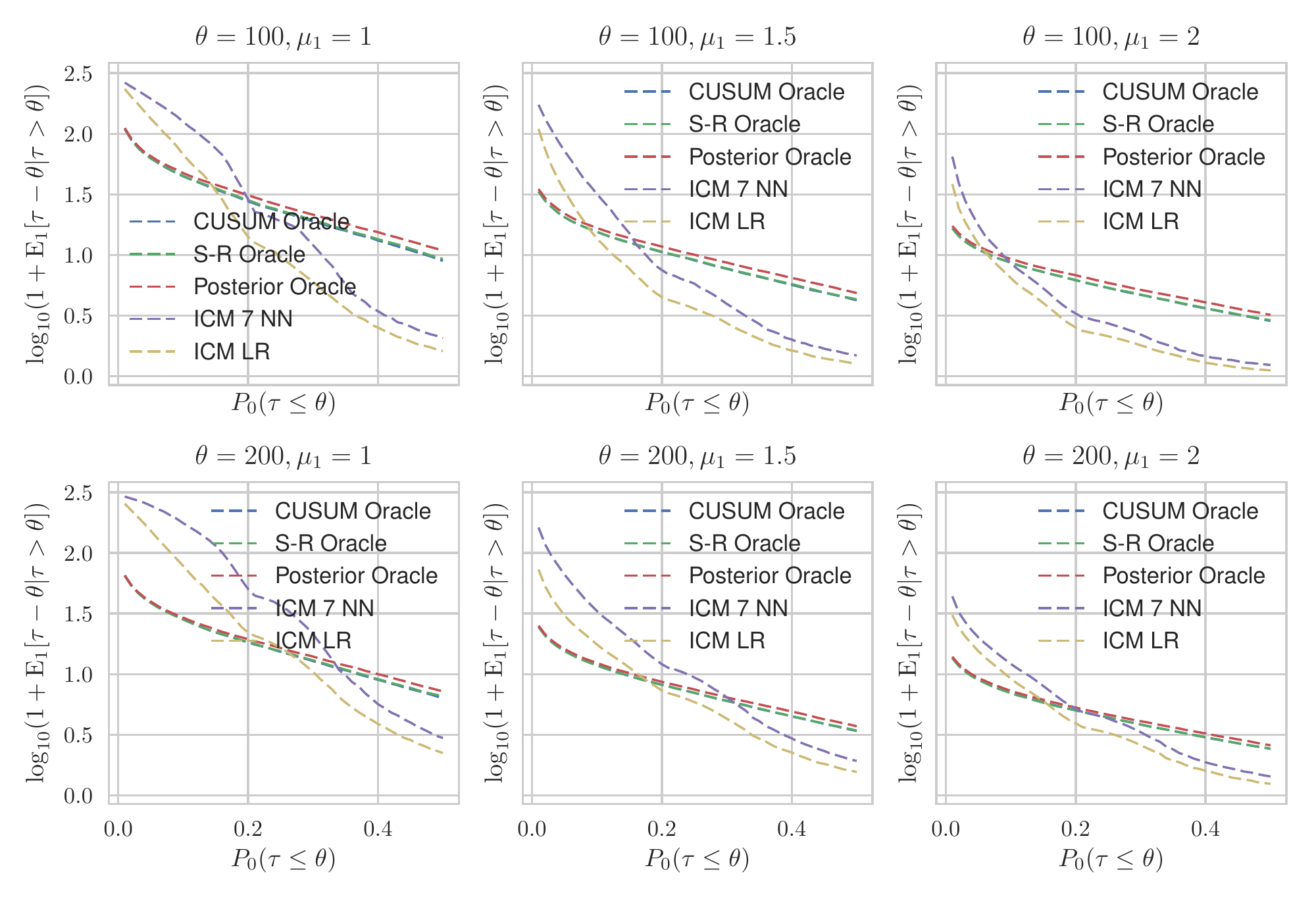}
    \caption{Mixture Betting Function}
    \label{fig:mixture_betting_function}
\end{figure}

Results for Mixture Betting Function are in Fig.~\ref{fig:mixture_betting_function}. Mean delays for some values of false alarm probability are in Tab. \ref{tab:mixture_betting_function}.

\begin{table}[bt]
\centering
\caption{Comparison of ICM (Mixture Betting Function) with Oracle by Mean Delay for different False Alarm probabilities}
\label{tab:mixture_betting_function}
\resizebox{\columnwidth}{!}{%
\begin{tabular}{|c|c|c|c|c|c|c|c|c|c|c|}
\hline
\multicolumn{1}{|l|}{\multirow{2}{*}{\textbf{Param.\textbackslash Probab. of FA}}} & \multicolumn{5}{c|}{\textbf{5\%}}                                                                                                                                                                                                                                                     & \multicolumn{5}{c|}{\textbf{10\%}}                                                                                                                                                                                                                                                    \\ \cline{2-11} 
\multicolumn{1}{|l|}{}                                                             & \textbf{ICM LR}  & \textbf{ICM kNN} & \textbf{\begin{tabular}[c]{@{}c@{}}CUSUM\\ Oracle\end{tabular}} & \multicolumn{1}{c|}{\textbf{\begin{tabular}[c]{@{}c@{}}S-R\\ Oracle\end{tabular}}} & \multicolumn{1}{c|}{\textbf{\begin{tabular}[c]{@{}c@{}}Posterior\\ Oracle\end{tabular}}} & \textbf{ICM LR}  & \textbf{ICM kNN} & \textbf{\begin{tabular}[c]{@{}c@{}}CUSUM\\ Oracle\end{tabular}} & \multicolumn{1}{c|}{\textbf{\begin{tabular}[c]{@{}c@{}}S-R\\ Oracle\end{tabular}}} & \multicolumn{1}{c|}{\textbf{\begin{tabular}[c]{@{}c@{}}Posterior\\ Oracle\end{tabular}}} \\ \hline
$\theta=100, \mu_1=1$ & 132.58 &  193.27 &  61.59 & 62.01 &  64.37 &  66.34 &  124.34 &  43.53 & 43.89 &  46.40 \\ \hline
$\theta=100, \mu_1=1.5$ & 32.73 &  71.01 &  19.51 & 19.51 &  20.98 &  12.63 &  30.77 &  14.50 & 14.51 &  15.67 \\ \hline
$\theta=100, \mu_1=2$ & 11.37 &  16.60 &  10.11 & 10.09 &  10.78 &  5.45 &  7.57 &  7.64 & 7.64 &  8.27 \\ \hline
$\theta=200, \mu_1=1$ & 151.61 &  244.65 &  37.78 & 37.80 &  38.73 &  77.10 &  175.08 &  27.24 & 27.24 &  28.25 \\ \hline
$\theta=200, \mu_1=1.5$ & 29.50 &  65.29 &  14.62 & 14.52 &  15.16 &  16.56 &  32.13 &  10.85 & 10.81 &  11.36 \\ \hline
$\theta=200, \mu_1=2$ & 14.49 &  19.12 &  8.02 & 7.98 &  8.30 &  8.20 &  11.16 &  6.00 & 5.97 &  6.28 \\ \hline
\end{tabular}
}
\end{table}

\subsection{Kernel Betting Function}

\begin{figure}[bt]
    \centering
    \includegraphics[width=16cm]{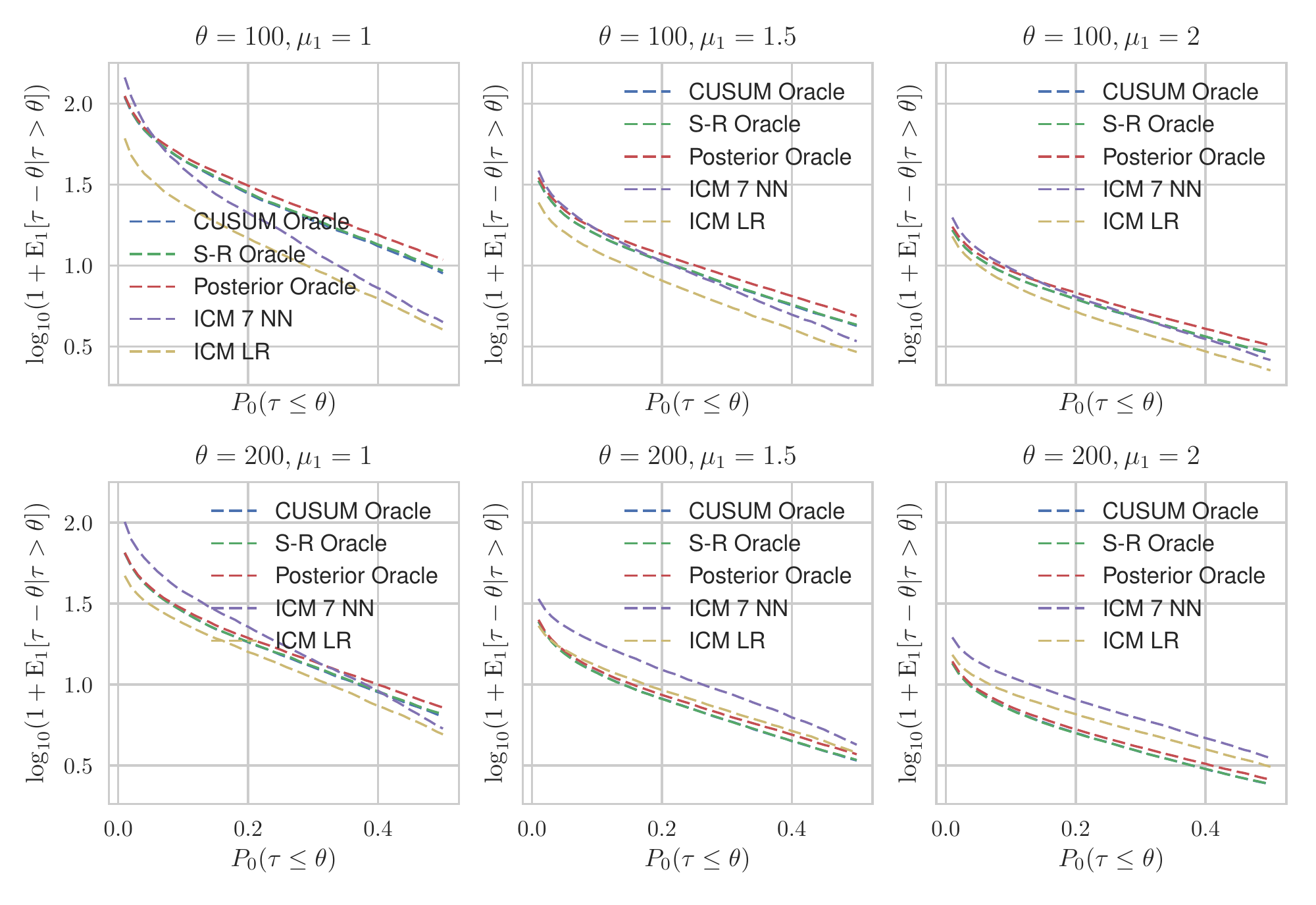}
    \caption{Kernel Density Betting Function}
    \label{fig:kernel_betting_function}
\end{figure}

Results for Kernel Betting Function are in Fig.\ \ref{fig:kernel_betting_function}. Mean delays for some values of false alarm probability are in Tab. \ref{tab:kernel_betting_function}. We use a sliding window of size $L=100$ to estimate density of p-values.

We can see, that results for the Kernel Betting Function is worse than for the Mixture Betting Function. The main reason is that it takes a long time for the martingale to grow sufficiently. In fact, before the change-point the distribution of p-values is uniform on $[0, 1]$. If for the current moment of time $n$ it holds that  $n-L\geq\theta$, the distribution of p-values $p_s$, $s\in[n-L,n]$ is also uniform. Thus, 
the martingale grows only when the change-point $\theta$ is inside the interval $[n-L,n]$, p-values from which are used for density estimation. This is the reason why in section \ref{sec:kernel_bet_fun} we propose new Precomputed Kernel Density Betting Function.

\begin{table}[bt]
\centering
\caption{Comparison of ICM (Kernel Density Betting Function) with Oracle by Mean Delay for different False Alarm probabilities}
\label{tab:kernel_betting_function}
\resizebox{\columnwidth}{!}{%
\begin{tabular}{|c|c|c|c|c|c|c|c|c|c|c|}
\hline
\multicolumn{1}{|l|}{\multirow{2}{*}{\textbf{Param.\textbackslash Probab. of FA}}} & \multicolumn{5}{c|}{\textbf{5\%}}                                                                                                                                                                                                                                                     & \multicolumn{5}{c|}{\textbf{10\%}}                                                                                                                                                                                                                                                    \\ \cline{2-11} 
\multicolumn{1}{|l|}{}                                                             & \textbf{ICM LR}  & \textbf{ICM kNN} & \textbf{\begin{tabular}[c]{@{}c@{}}CUSUM\\ Oracle\end{tabular}} & \multicolumn{1}{c|}{\textbf{\begin{tabular}[c]{@{}c@{}}S-R\\ Oracle\end{tabular}}} & \multicolumn{1}{c|}{\textbf{\begin{tabular}[c]{@{}c@{}}Posterior\\ Oracle\end{tabular}}} & \textbf{ICM LR}  & \textbf{ICM kNN} & \textbf{\begin{tabular}[c]{@{}c@{}}CUSUM\\ Oracle\end{tabular}} & \multicolumn{1}{c|}{\textbf{\begin{tabular}[c]{@{}c@{}}S-R\\ Oracle\end{tabular}}} & \multicolumn{1}{c|}{\textbf{\begin{tabular}[c]{@{}c@{}}Posterior\\ Oracle\end{tabular}}} \\ \hline
$\theta=100, \mu_1=1$ & 33.10 &  65.26 &  61.59 & 62.01 &  64.37 &  22.92 &  38.70 &  43.53 & 43.89 &  46.40 \\ \hline
$\theta=100, \mu_1=1.5$ & 15.08 &  22.03 &  19.51 & 19.51 &  20.98 &  11.15 &  15.65 &  14.50 & 14.51 &  15.67 \\ \hline
$\theta=100, \mu_1=2$ & 9.04 &  11.62 &  10.11 & 10.09 &  10.78 &  6.66 &  8.55 &  7.64 & 7.64 &  8.27 \\ \hline
$\theta=200, \mu_1=1$ & 30.06 &  54.14 &  37.78 & 37.80 &  38.73 &  22.90 &  36.57 &  27.24 & 27.24 &  28.25 \\ \hline
$\theta=200, \mu_1=1.5$ & 15.44 &  22.02 &  14.62 & 14.52 &  15.16 &  12.08 &  17.13 &  10.85 & 10.81 &  11.36 \\ \hline
$\theta=200, \mu_1=2$ & 10.00 &  12.81 &  8.02 & 7.98 &  8.30 &  7.83 &  10.15 &  6.00 & 5.97 &  6.28 \\ \hline
\end{tabular}
}
\end{table}

\subsection{Precomputed Kernel Betting Function}
When learning the Precomputed Kernel Betting Function we use one realization $z_1,\ldots,z_n,\ldots$ of length $1000$ with a CP at $\theta = 500$, such that $z_n\sim\mathcal{N}(\cdot\mid 0,1)$ for $n<\theta$ and $z_n\sim\mathcal{N}(\cdot\mid 1,1)$ for $n\geq\theta$ regardless of where the real CP is located and which amplitude it has.

\begin{figure}[bt]
    \centering
    \includegraphics[width=16cm]{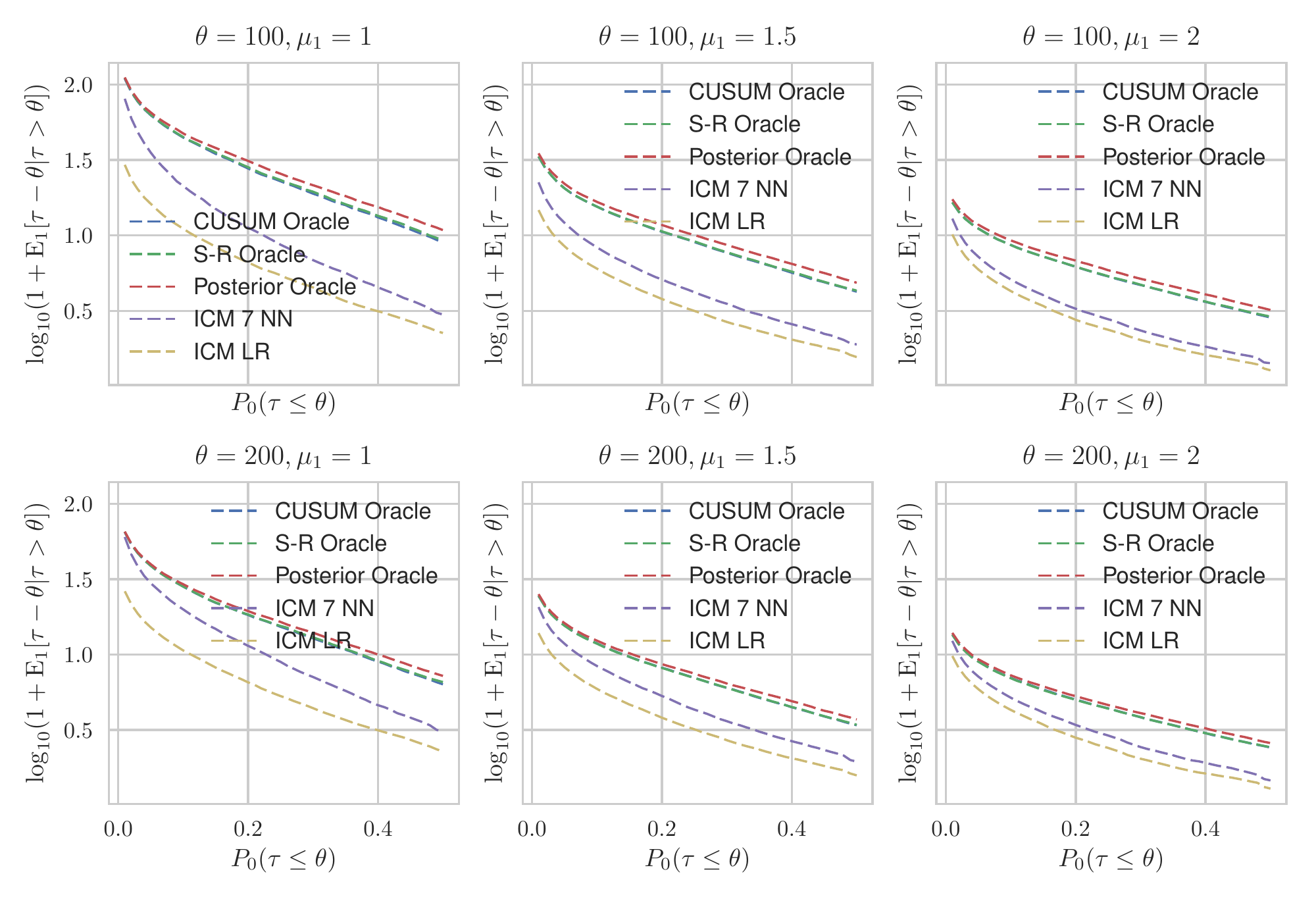}
    \caption{Precomputed Kernel Density Betting Function}
    \label{fig:oracle_precomputed_kernel_betting_function}
\end{figure}

Results for Precomputed Kernel Betting Function are in Fig.\ \ref{fig:oracle_precomputed_kernel_betting_function}. Mean delays for some values of false alarm probability are in Tab. \ref{tab:oracle_precomputed_kernel_betting_function}.

\begin{table}[bt]
\centering
\caption{Comparison of ICM (Precomputed Kernel Density Betting Function) with Oracle by Mean Delay for different False Alarm probabilities}
\label{tab:oracle_precomputed_kernel_betting_function}
\resizebox{\columnwidth}{!}{%
\begin{tabular}{|c|c|c|c|c|c|c|c|c|c|c|}
\hline
\multicolumn{1}{|l|}{\multirow{2}{*}{\textbf{Param.\textbackslash Probab. of FA}}} & \multicolumn{5}{c|}{\textbf{5\%}}                                                                                                                                                                                                                                                     & \multicolumn{5}{c|}{\textbf{10\%}}                                                                                                                                                                                                                                                    \\ \cline{2-11} 
\multicolumn{1}{|l|}{}                                                             & \textbf{ICM LR}  & \textbf{ICM kNN} & \textbf{\begin{tabular}[c]{@{}c@{}}CUSUM\\ Oracle\end{tabular}} & \multicolumn{1}{c|}{\textbf{\begin{tabular}[c]{@{}c@{}}S-R\\ Oracle\end{tabular}}} & \multicolumn{1}{c|}{\textbf{\begin{tabular}[c]{@{}c@{}}Posterior\\ Oracle\end{tabular}}} & \textbf{ICM LR}  & \textbf{ICM kNN} & \textbf{\begin{tabular}[c]{@{}c@{}}CUSUM\\ Oracle\end{tabular}} & \multicolumn{1}{c|}{\textbf{\begin{tabular}[c]{@{}c@{}}S-R\\ Oracle\end{tabular}}} & \multicolumn{1}{c|}{\textbf{\begin{tabular}[c]{@{}c@{}}Posterior\\ Oracle\end{tabular}}} \\ \hline
$\theta=100, \mu_1=1$ & 15.20 &  34.41 &  61.59 & 62.01 &  64.37 &  10.08 &  20.27 &  43.53 & 43.89 &  46.40 \\ \hline
$\theta=100, \mu_1=1.5$ & 7.47 &  11.12 &  19.51 & 19.51 &  20.98 &  5.02 &  7.32 &  14.50 & 14.51 &  15.67 \\ \hline
$\theta=100, \mu_1=2$ & 4.95 &  6.22 &  10.11 & 10.09 &  10.78 &  3.28 &  4.11 &  7.64 & 7.64 &  8.27 \\ \hline
$\theta=200, \mu_1=1$ & 14.14 &  28.70 &  37.78 & 37.80 &  38.73 &  9.65 &  18.91 &  27.24 & 27.24 &  28.25 \\ \hline
$\theta=200, \mu_1=1.5$ & 7.24 &  10.80 &  14.62 & 14.52 &  15.16 &  4.92 &  7.39 &  10.85 & 10.81 &  11.36 \\ \hline
$\theta=200, \mu_1=2$ & 4.90 &  6.15 &  8.02 & 7.98 &  8.30 &  3.29 &  4.18 &  6.00 & 5.97 &  6.28 \\ \hline
\end{tabular}
}
\end{table}

\subsection{Comparison with Optimal detectors}

\begin{figure}[bt]
    \centering
    \includegraphics[width=16cm]{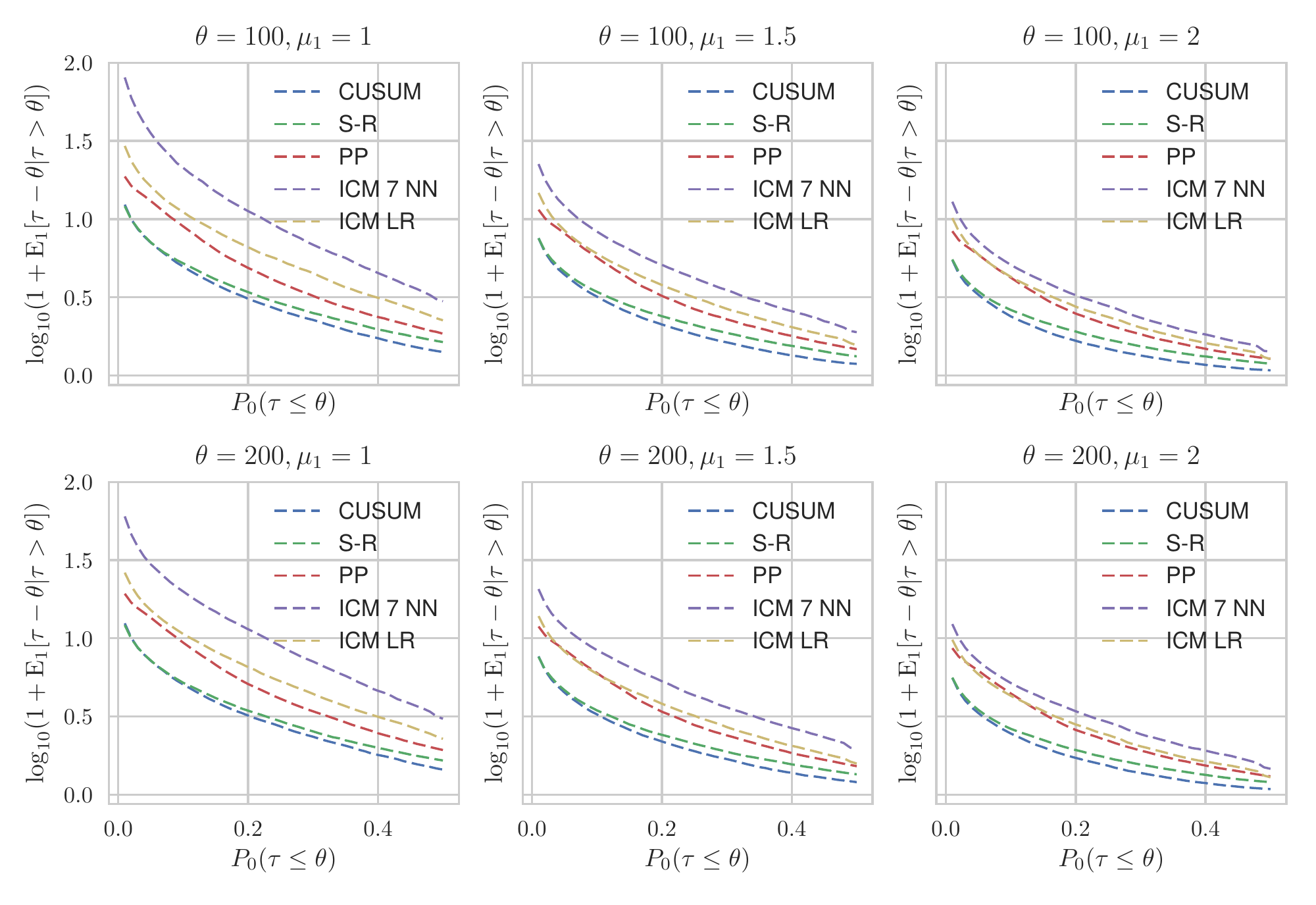}
    \caption{Comparison with Optimal detectors. Precomputed Kernel Density Betting Function}
    \label{fig:cusum_precomputed_kernel_betting_function}
\end{figure}

We also compare CP detection based on CMs with optimal detectors: Cumulative Sum (CUSUM), Shiryaev-Roberts (S-R) and Posterior Probability statistics (PP). One can see from Tab. \ref{tab:cusum_precomputed_kernel_betting_function} and Fig.\ \ref{fig:cusum_precomputed_kernel_betting_function} that our results are comparable to results of the optimal methods. CMs perform a little bit worse, but we should notice that it requires fewer assumptions (does not know the true $f_0$ and $f_1$) and is more general (distribution-free).

\begin{table}[bt]
\centering
\caption{Comparison of ICM (Precomputed Kernel Density Betting Function) with Optimal Detectors by Mean Delay for different False Alarm probabilities}
\label{tab:cusum_precomputed_kernel_betting_function}
\resizebox{\columnwidth}{!}{%
\begin{tabular}{|c|c|c|c|c|c|c|c|c|c|c|}
\hline
\multicolumn{1}{|l|}{\multirow{2}{*}{\textbf{Param.\textbackslash Probab. of FA}}} & \multicolumn{5}{c|}{\textbf{5\%}}                                                                                                                                                                                                                                                     & \multicolumn{5}{c|}{\textbf{10\%}}                                                                                                                                                                                                                                                    \\ \cline{2-11} 
\multicolumn{1}{|l|}{}                                                             & \textbf{ICM LR}  & \textbf{ICM kNN} & \textbf{\begin{tabular}[c]{@{}c@{}}CUSUM\\\end{tabular}} & \multicolumn{1}{c|}{\textbf{\begin{tabular}[c]{@{}c@{}}S-R\\ \end{tabular}}} & \multicolumn{1}{c|}{\textbf{\begin{tabular}[c]{@{}c@{}}Posterior\\ Prob.\end{tabular}}} & \textbf{ICM LR}  & \textbf{ICM kNN} & \textbf{\begin{tabular}[c]{@{}c@{}}CUSUM\\\end{tabular}} & \multicolumn{1}{c|}{\textbf{\begin{tabular}[c]{@{}c@{}}S-R\\\end{tabular}}} & \multicolumn{1}{c|}{\textbf{\begin{tabular}[c]{@{}c@{}}Posterior\\ Prob.\end{tabular}}} \\ \hline
$\theta=100, \mu_1=1$ & 15.20 &  34.41 &  6.08 & 6.11 &  12.06 &  10.08 &  20.27 &  3.97 & 4.22 &  7.99 \\ \hline
$\theta=100, \mu_1=1.5$ & 7.47 &  11.12 &  3.42 & 3.60 &  7.11 &  5.02 &  7.32 &  2.19 & 2.43 &  4.67 \\ \hline
$\theta=100, \mu_1=2$ & 4.95 &  6.22 &  2.29 & 2.46 &  4.93 &  3.28 &  4.11 &  1.39 & 1.63 &  3.23 \\ \hline
$\theta=200, \mu_1=1$ & 14.14 &  28.70 &  6.19 & 6.22 &  12.55 &  9.65 &  18.91 &  4.07 & 4.19 &  8.38 \\ \hline
$\theta=200, \mu_1=1.5$ & 7.24 &  10.80 &  3.50 & 3.66 &  7.44 &  4.92 &  7.39 &  2.26 & 2.46 &  4.99 \\ \hline
$\theta=200, \mu_1=2$ & 4.90 &  6.15 &  2.33 & 2.48 &  5.22 &  3.29 &  4.18 &  1.46 & 1.64 &  3.44 \\ \hline
\end{tabular}
}
\end{table}

\section{Conclusion}
\label{sec6}
In this paper we describe an adaptation of Conformal Martingales for change-point detection problem. We demonstrate the efficiency of this approach by comparing it with natural oracles, which are likelihood-based change-point detectors. Our results indicate that the efficiency of change-point detection based on conformal martingales in most of cases is comparable with that of oracle detectors. 

We propose and compare several approaches to calculating a betting function (a function that transforms p-values into a martingale) and a non-conformity measure (a function that defines strangeness and, therefore, p-values). We get that the Precomputed Kernel Betting Function provides the most efficient results and the Mixture Betting Function provides the worst results.

We also compare Inductive Conformal Martingales with methods that are optimal for known pre- and post-CP distributions, such as CUSUM, Shiryaev-Roberts and Posterior Probability statistics. Our results are  worse but still they are comparable. Some deterioration is inevitable, of course, since CMs are distribution-free methods and, therefore, require much weaker assumptions.

\acks\begingroup
We are grateful for the support from the European Union's Horizon 2020 Research and Innovation programme under Grant Agreement no.\ 671555 (ExCAPE project). The research presented in Section \ref{sec5} of this paper was supported by the RFBR grants 16-01-00576 A and 16-29-09649 ofi\_m. This work was also supported by the Russian Science Foundation grant (project 14-50-00150), the UK EPSRC grant (EP/K033344/1), and the Technology Integrated Health Management (TIHM) project awarded to the School of Mathematics and Information Security at Royal Holloway.  We are indebted to Prof.\ Ilya Muchnik, School of Data Analysis, Yandex, and Royal Holloway, University of London, for the studentship support of one of the authors.
\endgroup

\bibliography{bibs}
\end{document}